\documentclass[times,preprint,twocolumn,authoryear]{elsarticle}

\usepackage{amsmath,amsfonts}
\usepackage{array}
\usepackage{stfloats}
\usepackage{verbatim}
\usepackage{graphicx}
\usepackage{siunitx}
\usepackage{physics}
\usepackage{multirow}
\usepackage{mathtools}
\usepackage{utfsym}
\usepackage{booktabs}
\usepackage{caption}
\usepackage{subcaption}
\usepackage{comment}
\usepackage[capitalize]{cleveref}
\Crefname{figure}{Fig.}{Figs.}

\usepackage{ycviu}
\usepackage{framed,multirow}

\usepackage{amssymb}
\usepackage{latexsym}

\usepackage{url}
\usepackage{xcolor}
\definecolor{newcolor}{rgb}{.8,.349,.1}

\DeclarePairedDelimiter{\floor}\lfloor\rfloor

\makeatletter
\DeclareRobustCommand\onedot{\futurelet\@let@token\@onedot}
\def\@onedot{\ifx\@let@token.\else.\null\fi\xspace}

\def\ie{\emph{i.e}\onedot}

\makeatother

\renewcommand{\ie}{\emph{i.e.}}

\journal{****}

\begin{document}
\ifpreprint
  \setcounter{page}{1}
\else
  \setcounter{page}{1}
\fi

\begin{figure*}
\begin{minipage}{\linewidth}
\begin{frontmatter}

\title{A Single Set of Adversarial Clothes Breaks Multiple Defense Methods in the Physical World}

\author[thu_cst]{Wei Zhang}
\author[berkeley]{Zhanhao Hu}
\author[thu_cst]{Xiao Li}
\author[thu_cst]{Xiaopei Zhu}
\author[thu_cst,cibr]{Xiaolin Hu\corref{cor1}}
\cortext[cor1]{Corresponding author: 
}
\ead{xlhu@tsinghua.edu.cn}

\address[thu_cst]{Department of Computer Science and Technology, Institute for Artificial Intelligence, THBI, BNRist, Tsinghua University, Beijing 100084, China}
\address[berkeley]{Berkeley Institute for Data Science, Department of Electrical Engineering and Computer Sciences, University of California, Berkeley, CA 94720, USA}
\address[cibr]{Chinese Institute for Brain Research (CIBR), Beijing 100010, China}

\received{***}
\finalform{***}
\accepted{***}
\availableonline{***}
\communicated{***}

\begin{abstract}
In recent years, adversarial attacks against deep learning-based object detectors in the physical world have attracted much attention. To defend against these attacks, researchers have proposed various defense methods against adversarial patches, a typical form of physically-realizable attack.
However, our experiments showed that simply enlarging the patch size could make these defense methods fail.
Motivated by this, we evaluated various defense methods against adversarial clothes which have large coverage over the human body. Adversarial clothes provide a good test case for adversarial defense against patch-based attacks because they not only have large sizes but also look more natural than a large patch on humans.
Experiments show that all the defense methods had poor performance against adversarial clothes in both the digital world and the physical world.
In addition, we crafted a single set of clothes that broke multiple defense methods on Faster R-CNN. The set achieved an Attack Success Rate (ASR) of \SI{96.06}{\percent} against the undefended detector and over \SI{64.84}{\percent} ASRs against nine defended models in the physical world, unveiling the common vulnerability of existing adversarial defense methods against adversarial clothes. Code is available at: \url{https://github.com/weiz0823/adv-clothes-break-multiple-defenses}.
\end{abstract}

\begin{keyword}
\KWD Deep learning security \sep
adversarial example \sep
adversarial texture \sep
adversarial robustness \sep
model vulnerability
\end{keyword}

\end{frontmatter}
\end{minipage}
\end{figure*}

\section{Introduction}
\label{sec:intro}

Deep Neural Networks (DNNs) are known to be vulnerable to adversarial examples
not only in the digital world \citep{goodfellow2014fgsm,madry2017pgd,lavan}, but also in the physical world \citep{adversarial-patch,thys2019fooling,xu2020adv-tshirt,wu2020adv-cloak,hu2021nat-patch,hu2022adv-texture,hu2023camou}.
Physical adversarial examples raise serious security concerns since they can be deployed in the real world. Given the widespread deployment of object detection models in various applications, researchers have focused on fooling object detection models in the physical world in recent years, especially person detection models \citep{thys2019fooling,xu2020adv-tshirt,wu2020adv-cloak,hu2021nat-patch,hu2022adv-texture,hu2023camou}.

To defend against physically realizable attacks, various defense methods \citep{lgs-naseer2019,epgf-zhou2020,udf-yu2022,robust-self-attention,adv-yolo-ji2021,fnc-yu2021,ape-kim2022,sac-liu2022,over-activation,patchzero,jedi-tarchoun2023,rao2020patch-at,metzen2021meta-at} have been proposed. They are usually designed for and evaluated on adversarial patches \citep{adversarial-patch,thys2019fooling}.
Some of the defense methods \citep{udf-yu2022,ape-kim2022,epgf-zhou2020} were evaluated on physical-world patch-based attacks, and others were evaluated on digital-world patch-based attacks. They worked well in defending against patch-based attacks.
However, in this paper we will show that this may have given a false sense of security of protected object detectors.

\begin{figure}[t]
    \centering
    \includegraphics[width=0.8\linewidth]{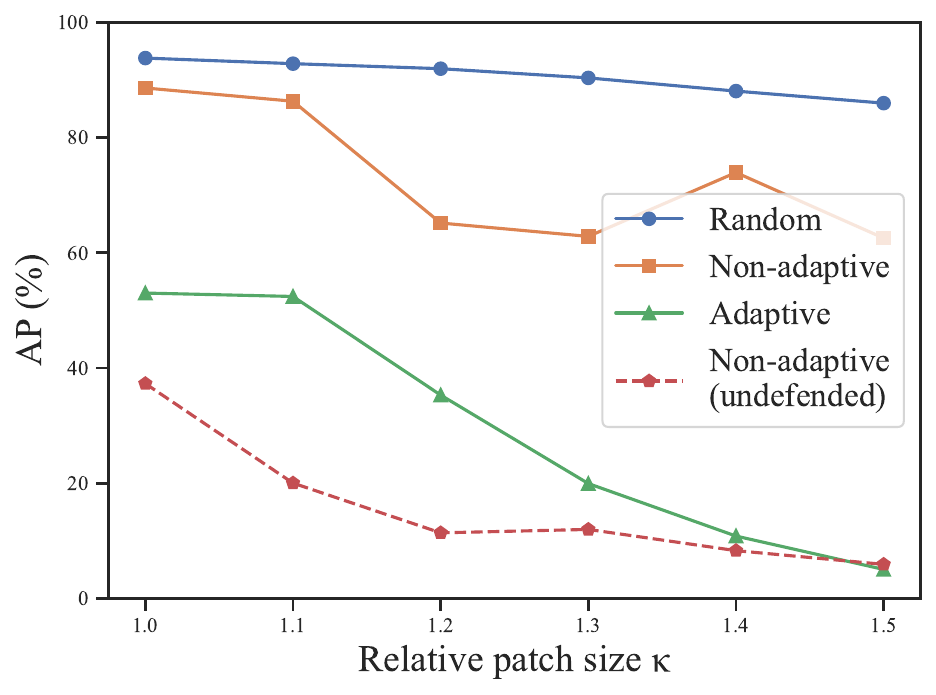}
    \caption{APs of \textit{person} class of the FNC-defended Faster R-CNN \citep{frcnn} detector against patch attack of different patch sizes on the Inria person dataset \citep{inria}.
    Solid lines denote the APs of the FNC-defended detector, when the input images are applied with \textit{random patch}, \textit{non-adaptive patch} optimized on the undefended detector, and \textit{adaptive patch} optimized on the FNC-defended detector. Dashed line denotes the AP of the undefended detector, when the input images are applied with \textit{non-adaptive patch}. Details of experiment setups are provided in \cref{sec:expand-patch}.
    }
    \label{fig:area_intro}
\end{figure}

Our key observation is that existing defense methods against patch-based attacks have not given enough attention to the patch size.
Intuitively, larger patch size corresponds to larger optimization space, and the attack becomes stronger and harder to defend against.
Motivated by this, we first investigated whether patch size influences the performance of the defense. \Cref{fig:area_intro} provides a clue, where we tested FNC \citep{fnc-yu2021}, a patch defense method that was independent of the patch size.
Note that the defense performance of FNC against adaptive attack (the green line) vanished as the patch became larger.
However, simply enlarging patch is not a natural way for physical implementation, since holding a board much larger than body size is not very much natural in real-world scenarios.
Instead, texture-based attacks \citep{hu2022adv-texture,hu2023camou}, also known as adversarial clothes, provide intuitively similar implementation as the enlarged patch size.
It is therefore necessary to re-evaluate the state-of-the-art (SOTA) defense methods to demonstrate their capability of defending against adversarial clothes.

In this study, we evaluated various SOTA adversarial defense methods against adversarial clothes.
To craft the adversarial clothes in the physical world, we need to design the texture by gradient-based optimization in the digital world first.
We adopted a 3D rendering pipeline \citep{hu2023camou} to optimize adversarial textures on clothes.
We tested diverse defense methods \citep{li2023od-at,fnc-yu2021,epgf-zhou2020,sac-liu2022,ape-kim2022,lgs-naseer2019,udf-yu2022,jedi-tarchoun2023} and found that all of them had poor performance against the digital-world adversarial texture.
An implementation of the adversarial clothes in the physical world by printing the texture on real-world clothes unveiled that the vulnerability of defense methods could also be exploited in real-world scenarios.

Based on these findings, we conjectured that these defense methods share common vulnerabilities that could be exploited by a single set of adversarial clothes.
We optimized the texture of a set of clothes and included an ensemble of defended models during the optimization.
We printed the clothing texture on a piece of cloth and tailored it into a set of adversarial clothes including a shirt and a pair of trousers.
Experiments showed that the set of clothes bypassed nine defense methods in both the digital world and the physical world.

The main contributions are:
(1) We evaluated SOTA adversarial defense methods against adversarial clothes, and found that adversarial clothes impaired the performance of the existing defense methods in real-world scenarios.
(2) We successfully broke nine defense methods in the physical world with a single set of clothes, achieving over \SI{64.84}{\percent} ASRs against all nine defense methods.
(3) The results unveil that SOTA defense methods are still vulnerable to physical-world adversarial examples when confronted with texture-based attacks.

The rest of this paper is organized as follows.
\Cref{sec:settings} briefly reviews the physically realizable threats to person detectors.
\Cref{sec:expand-patch} shows the impact of patch size on adversarial defense, which provides the motivation to evaluate the defense methods against adversarial clothes.
\Cref{sec:method} describes the attack settings of adversarial clothes and evaluation metrics used in this study.
\Cref{sec:adaptive-3d} shows the evaluation results of various adversarial defense methods against adversarial clothes with both the digital-world results and the physical-world results.
Finally, the conclusion is given in \cref{sec:conclusion}.

\section{Related work}
\label{sec:settings}

In this section, we introduce the threats including patch-based attacks and texture-based attacks to fool person detectors.
Then, we briefly review the defense methods against physically realizable attacks.

\subsection{Threats}

\paragraph{Adversarial Patches}
The first work for generating physically realizable adversarial patches to fool person detectors was proposed by \citet{thys2019fooling},
and the similar pipeline has been followed up by several works \citep{xu2020adv-tshirt,wu2020adv-cloak,hu2021nat-patch}.
We denote their method as \emph{AdvPatch}.
The main process is to transform and apply an image patch onto each person in an image from the training dataset according to the person's bounding box, and
then optimize all the patches by minimizing the detection scores outputted by the target detector.
The loss is defined as the maximum detection score among all bounding boxes in each image, which we minimize.

To make the patches smoother, the AdvPatch method adds a total variation (TV) loss.
TV loss is lower when neighboring pixel values are closer, and the patch has smoother appearance.
There is also a non-printability score (NPS) loss term to restrict pixels in the patches within a set of printable colors.
In addition, Expectation over Transformations (EoT) \citep{athalye2018eot} is applied to the patches to make them more robust to physical transformations, including randomizing locations, rotations, brightness, contrast, and pixel noises.

\citet{hu2022adv-texture} extended the adversarial patches to tileable patches in order to make the attack effective in multiple viewing angles.
The physical-world implementation is made by printing tileable patch repeatedly as texture on a piece of cloth, then tailoring the cloth into clothes covered with adversarial patterns.
We denote the method as \textit{AdvTexture}.
Despite the notation, the optimization pipeline of the attack is still based on the adversarial patches, while employing a toroidal cropping technique to randomly crop a unit of the patch, and expandable generation technique to generate the tileable patch with generation model.

\paragraph{Texture-based attacks}

\label{sec:setting-3d}

\citet{hu2023camou} proposed to use a 3D rendering pipeline to obtain adversarial camouflage texture (AdvCaT) for clothes. The AdvCaT resembles typical camouflage patterns, making the clothes natural-looking. During optimization, a 3D person mesh model is rendered from different viewing angles and the rendered images are synthesize with background images. In addition to EoT used for adversarial patches, Thin Plate Spline (TPS) \citep{bookstein1989tps,donato2002approx-tps} deformation is also incorporated to enhance the robustness of the attack in the physical world.

\subsection{Defenses}
\label{sec:related-defenses}
The defense methods against physically realizable attacks can be roughly divided into four categories, we briefly review them as follows.

\paragraph{Model-independent input preprocess}
This kind of defense methods \citep{lgs-naseer2019,epgf-zhou2020,jedi-tarchoun2023,sac-liu2022,patchzero,jing2024pad} either mask out or suppress the regions on the input images that are suspected to contain adversarial patches, before the input images are fed into the detector.
Specifically, Local Gradient Smoothing (LGS) \citep{lgs-naseer2019} computes the gradients of the pixels in an image with respect to pixel position, then suppresses large-gradient regions by a factor proportional to that gradient computed.
Entropy-based methods \citep{epgf-zhou2020,jedi-tarchoun2023} compute the entropy \citep{entropy} of the pixels within a sliding window across the input image. High-entropy regions are suspected as adversarial regions. Information Distribution Based Defense (IDBD) \citep{epgf-zhou2020} incorporates entropy-based proposal with gradient-based filtering which computes the sum of the pixel gradients within the sliding window. 
Jedi \citep{jedi-tarchoun2023} localizes adversarial patches with entropy heatmap, then completes the patch mask with an autoencoder, and finally inpaints the detected adversarial patch region.
PAD \citep{jing2024pad} localizes adversarial patches with mutual information score and compression difference. However, because the computational cost of mutual information is squared to the cost of computing entropy on a sliding window, the processing speed of PAD is very slow.
Segment and Complete (SAC) \citep{sac-liu2022} trains a patch segmentation model that outputs a raw mask indicating the regions of the patches. A shape completion algorithm is then applied to the raw mask, generating a completed patch mask. Finally, the patch region is removed based on the completed patch mask.
Similarly, NAPGuard \citep{wu2024napguard} trains a patch detection model that localizes the bounding box of patch.

\paragraph{Outlier feature filter}
Defense methods in this category \citep{fnc-yu2021,ape-kim2022,over-activation,robust-self-attention} extract inner features from the target DNN model. They usually filter or clip the feature vectors according to their distributions.
Feature Norm Clip (FNC) \citep{fnc-yu2021} is motivated by the observation that the $l_2$ norms of the convolutional neural network (CNN) feature vectors at the regions containing adversarial patches are usually larger than those of the benign regions. All feature maps of the CNN models are filtered with a clip operation, making an upper bound on the norm of feature vectors.
Adversarial Patch-feature Energy (APE) \citep{ape-kim2022} combines adversarial region detection with feature filtering. Adversarial regions are detected based on multi-level outlier features. Then the first-layer outlier features are clipped within the detected adversarial regions.

\paragraph{Adversarial training}
Adversarial Training (AT) \citep{madry2017pgd} and its variants \citep{zhang2019trades,wu2020awp,rock} are usually recognized as the most effective methods in defending classification models against noise-bounded adversarial attacks \citep{robustbench}.
Recently, AT has been applied to object detectors \citep{zhang2019od-at,chen2021class-aware-odat,dong2022adv-aware-odat,li2023od-at}, under the setting that the adversarial noise is bounded by $l_p$ norms.
However, as far as we know, no AT method has been proposed specifically for object detectors against physical attacks.
In experiments, we utilized the checkpoints of the AT models from \citet{li2023od-at}, which is one of the SOTA AT methods for object detection against $l_\infty$ norm bounded attack. The AT models were trained with the $l_\infty$ norm bound $\epsilon=\flatfrac{4}{255}$.

\paragraph{Defensive frame}
\citet{udf-yu2022} proposed to train a Universal Defensive Frame (UDF) on adversarial examples.
The method involves training the UDF in conjunction with the attacking patch, following a pipeline similar to AT.
\citet{mao2024udfilter} proposed a similar method that optimized the defense filter in conjunction with the attacking patch. The defense filter was defined as a noise image that was linearly interpolated with the input image to robustify the detection.

\section{Impact of patch size on adversarial defense}
\label{sec:expand-patch}

Previous adversarial patch attacks \citep{thys2019fooling,hu2021nat-patch,hu2022adv-texture} mainly focused on patches with a fixed size and scaled them proportional to the size of target bounding box.
This is the setting that previous defense methods defended against.
Intuitively, larger patches on the target should have better adversarial effect.
To the best of our knowledge, no prior studies have quantitatively assessed the extent of adversarial effects that larger patches can achieve, particularly in the context of defended models.
To show that the patch size does have impact on the performance of defense methods, we took FNC \citep{fnc-yu2021}, a typical size-independent defense method, as an example, and used AdvPatch \citep{thys2019fooling} as the attack method.
Faster R-CNN \citep{frcnn} was chosen as the target detector. We mainly studied the detection performance of the detector defended by the FNC method against the AdvPatch attack with both non-adaptive patches and FNC-adaptive patches.

\begin{figure}[t]
    \centering
    \includegraphics[width=\linewidth]{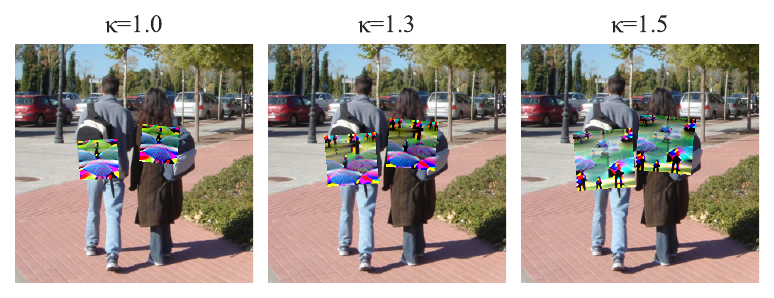}
    \caption{Visualization of the patches of different sizes when applied onto persons in the Inria \citep{inria} dataset.
    The images have been cropped and zoomed in for a better view.
    }
    \label{fig:vis-scale}
\end{figure}

For the patch-based attack, the size of the patch applied onto a person was determined by the diagonal length of the person's ground truth bounding box.
Suppose we have a bounding box of the target object with diagonal length $d$, the edge length $l$ of the patch can be calculated by $l=cd$, where $c$ is a constant controlling patch size.
We used $c_0=0.2$ as the constant $c$ for the baseline patch, consistent with the previous study \citep{thys2019fooling}.
To scale up the patch, we adjusted the value of $c$, and denote $\kappa=\flatfrac{c}{c_0}$ as \textit{relative patch size}.
We tested the defense performance of FNC under both non-adaptive and adaptive attacks with $\kappa\in[1.0, 1.5]$.
An visualization of patches in different sizes in the digital world is provided in \cref{fig:vis-scale}.
We used the same patch resolution, $300\times 300$ pixels, for different patch sizes, in order to keep the dimension of the attack solution space the same.
As shown in \cref{fig:area_intro}, the FNC-defended Faster R-CNN detector had APs larger than \SI{53.01}{\percent} against all types of patches with sizes of $\kappa=1.0$.
In comparison, the undefended Faster R-CNN detector had an AP of only \SI{37.30}{\percent}.
As $\kappa$ increased, the APs of the FNC-defended detector against non-adaptive patches remained high (over \SI{60}{\percent}), while the APs of of the detector against adaptive patches dropped quickly.
When $\kappa$ reached $1.5$, there was no significant difference between the AP of undefended detector and that of FNC-defended detector against the strongest patch (\ie the adaptive patch).
The results showed that the defense performance of FNC vanished as the patch became larger.

\section{Attack settings of adversarial clothes}
\label{sec:method}

As shown in \cref{sec:expand-patch}, expanding patch size does have an impact on defense performance. $\kappa=1.5$ made the defense performance of FNC vanish.
However, simply enlarging patch is not a natural way for physical implementation, since holding a board much larger than body size is not natural in real-world scenarios.
Instead, texture-based attacks \citep{hu2022adv-texture,hu2023camou}, also known as adversarial clothes, provide intuitively similar implementation as the enlarged patch size.
It is therefore necessary to re-evaluate the SOTA defense methods to demonstrate their capability of defending against adversarial clothes.

To systematically evaluate the defense models, we targeted nine typical defense methods, including AT \citep{li2023od-at}, FNC \citep{fnc-yu2021}, LGS \citep{lgs-naseer2019}, IDBD \citep{epgf-zhou2020},
SAC \citep{sac-liu2022}, APE \citep{ape-kim2022}, UDF \citep{udf-yu2022}, Jedi \citep{jedi-tarchoun2023}, and NAPGuard \citep{wu2024napguard}.
These defense methods cover all four categories of adversarial patch defenses as detailed in \cref{sec:related-defenses}.
The target detectors were Faster R-CNN \citep{frcnn} and FCOS \citep{fcos}, which represent typical two-stage and single-stage detectors, respectively.
Both detectors were pretrained on MS-COCO \citep{coco} dataset.
Input images were cropped or padded to equal width and height and resized to $416\times416$, then normalized before being fed into the detector.

\subsection{Optimization in the digital world}
\label{sec:method-digital}

We utilized the 3D rendering pipeline proposed by \citet{hu2023camou}.
Since the naturalness of clothing textures was not a concern in this study, we excluded the module for camouflage patterns generation in AdvCaT \citep{hu2023camou}, and optimized the texture map of the 3D mesh model pixel-wise.
This form of attack is denoted as \textit{Texture3D}.
To improve the robustness of physical implementation of the clothing textures,
we incorporated TV loss computed on the texture map during texture optimization.

We used the same background image dataset as used by AdvCaT \citep{hu2023camou},
consisting of 506 background images varying in the scene.
The background images were split into 376 images for training and 130 images for testing.
Textures were optimized for 100 epochs with Adam \citep{adam} optimizer.
The initial learning rate for Texture3D was set to $0.01$.

\subsection{Enhancing physical world robustness of the attack}
\begin{figure}[t]
    \centering
    \includegraphics[width=0.6\linewidth]{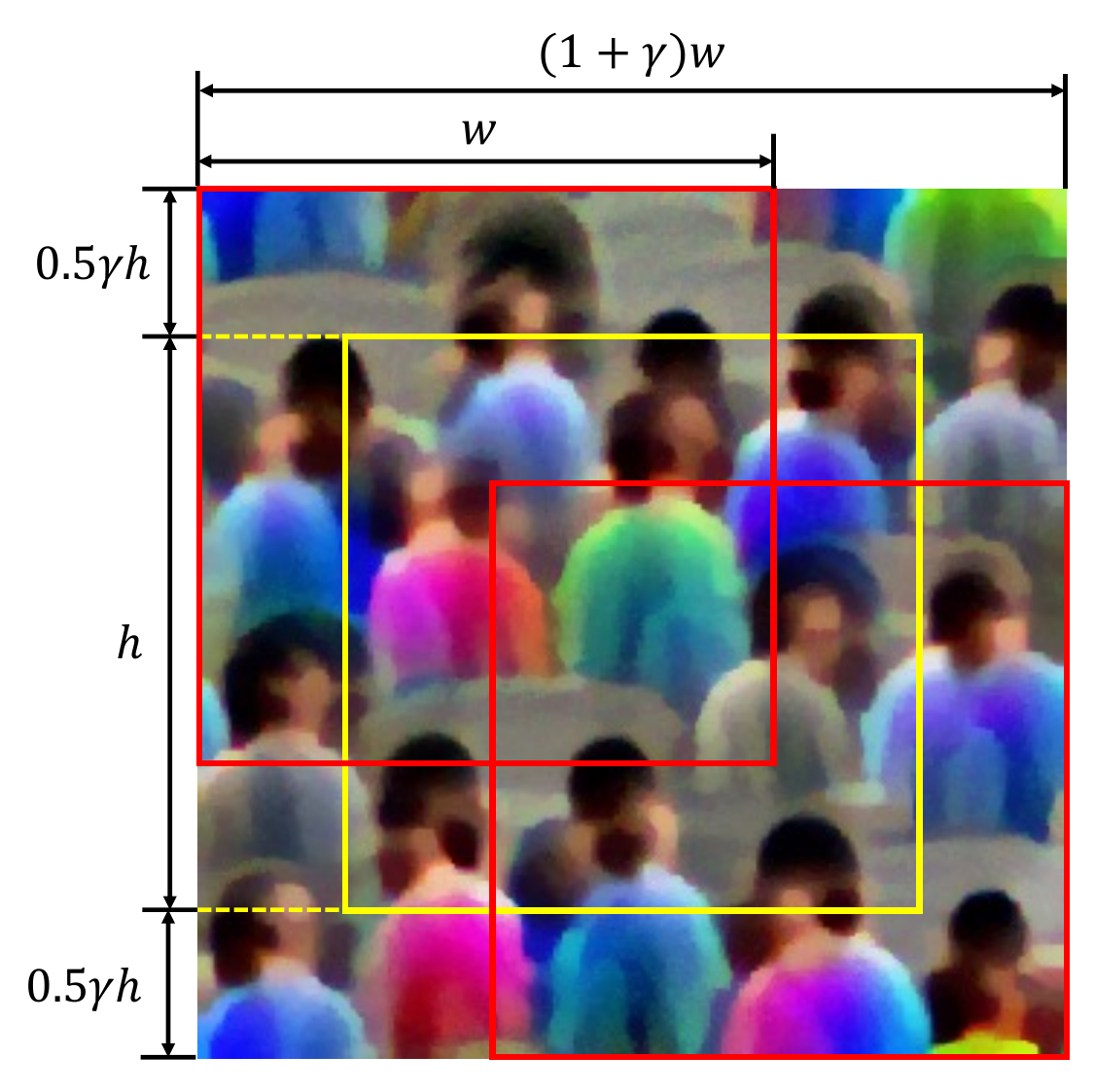}
    \caption{Illustration of position jittering crops to enhance the physical world robustness against texture-based attacks. The whole texture represents the latent feature map. The red boxes are possible texture crops used for optimization, and the yellow box is the center texture crop used for evaluation and physical implementation.}
    \label{fig:illust-jitter}
\end{figure}

Besides incorporating the EoTs to migrate the gap between digital-space optimization and physical-world implementation as described in the original 3D rendering attack pipeline \citep{hu2023camou}, we introduce \emph{position jittering} to the texture map to further simulate physical-world transformations, as shown in \cref{fig:illust-jitter}.
Suppose the size of the texture map is $(w,h)$, and the jittering intensity is $0<\gamma<1$.
Instead of optimizing the original texture map, 
we pad the texture map and optimize on the padded \emph{latent texture map} with size $(\floor{(1+\gamma) w},\floor{(1+\gamma) h})$, where $\floor{\cdot}$ denotes the rounding down operation.
For each optimization iteration, we crop a new texture map with fixed size $(w,h)$ and random top-left corner coordinate ranging from $(0,0)$ to $(\floor{\gamma w},\floor{\gamma h})$ from the latent texture map for rendering (see \cref{fig:illust-jitter}, red boxes).
During evaluation and for physical implementation, the texture map with size $(w,h)$ and top-left coordinate $(\floor{0.5\gamma w},\floor{0.5\gamma h})$ is used (see \cref{fig:illust-jitter}, yellow box).

\subsection{Physical world implementation}

Following \citet{hu2022adv-texture,hu2023camou}, we printed the texture map optimized by the texture-based attack on pieces of cloth and tailored them into shirts and trousers to produce sets of adversarial clothes.

\subsection{Evaluation}

\paragraph{Digital world evaluation}
For the evaluation metric in the digital world, we used Average Precision (AP) on the \textit{person} class.
The Intersection over Union (IoU) threshold was set to $0.5$ for both patch-based and texture-based attacks, consistent with previous works \citep{thys2019fooling,hu2021nat-patch,hu2022adv-texture,hu2023camou}.

\paragraph{Physical world evaluation}
To evaluate the effectiveness of the clothes in the physical world, we used the metric of Attack Success Rates (ASRs) on a set of images.
ASR calculates the percentage of successfully attacked images in all test images.
The attack was considered as successful if no box of \textit{person} class had an IoU over $0.5$.
In addition, only boxes with confidence scores larger than $0.5$ were taken into account.

Two actors (age mean: 25; age range: \numrange{22}{28};
height range: \SIrange{178}{188}{cm}) were recruited to collect physical test data. The physical test results were all averages of the results on the two subjects. The recruitment and study procedures were
approved by the Department of Psychology Ethics Committee, Tsinghua University, Beijing, China.

To evaluate the average attack performance from multiple viewing angles, we recorded a video of person turning circles,
and evenly extracted frames from the video to form our test set.
The ground truth bounding boxes were manually annotated.
These evaluation metrics are consistent with previous works \citep{hu2022adv-texture,hu2023camou}.

\section{Results}
\label{sec:adaptive-3d}

\begin{table}[t]
\small
\caption{APs (\si{\percent}, $\uparrow$) of Faster R-CNN equipped with different defenses (including the undefended detector) against texture-based attacks.
    AT and FNC defense methods were evaluated using adaptive attack (marked with $\dagger$).}
    \label{tab:dig-fr-3d}
    \centering
    \begin{tabular}{ccc}
        \toprule
        Model & Random & Adversarial \\
        \midrule
        Undefended & 88.52 & 0.03 \\
        AT \citep{li2023od-at} & 93.72 & $5.97^\dagger$ \\
        FNC \citep{fnc-yu2021} & 89.05 & $0.91^\dagger$ \\
        LGS \citep{lgs-naseer2019} & 89.21 & 0.87 \\
        IDBD \citep{epgf-zhou2020} & 88.31 & 1.72 \\
        SAC \citep{sac-liu2022} & 48.46 & 0.43 \\
        APE \citep{ape-kim2022} & 88.31 & 0.05 \\
        UDF \citep{udf-yu2022} & 73.25 & 1.51 \\
        Jedi \citep{jedi-tarchoun2023} & 88.65 & 5.32 \\
        NAPGuard \citep{wu2024napguard} & 87.35 & 0.08 \\
        \bottomrule
    \end{tabular}
\end{table}

\begin{table}[t]
\small
\caption{APs (\si{\percent}, $\uparrow$) of FCOS \citep{fcos} equipped with different defenses (including the undefended detector) against texture-based attacks.
    AT and FNC defense methods are evaluated using adaptive attack (marked with $\dagger$).}
    \label{tab:dig-fcos-3d}
    \centering
    \begin{tabular}{ccc}
        \toprule
        Model & Random & Adversarial \\
        \midrule
        Undefended & 63.67 & 0.05 \\
        AT \citep{li2023od-at} & 91.56 & $5.24^\dagger$ \\
        FNC \citep{fnc-yu2021} & 74.40 & $0.15^\dagger$ \\
        LGS \citep{lgs-naseer2019} & 61.90 & 0.10 \\
        IDBD \citep{epgf-zhou2020} & 64.11 & 0.05 \\
        SAC \citep{sac-liu2022} & 29.60 & 0.04 \\
        APE \citep{ape-kim2022} & 63.01 & 0.03 \\
        UDF \citep{udf-yu2022} & 76.15 & 6.74 \\
        Jedi \citep{jedi-tarchoun2023} & 61.38 & 0.06 \\
        NAPGuard \citep{wu2024napguard} & 62.46 & 0.02 \\
        \bottomrule
    \end{tabular}
\end{table}

We evaluated the performance of detectors equipped with various defense methods against the texture-based attack Texture3D.

\subsection{Digital-world evaluation}

\begin{table*}[t]
\small
\caption{APs (\si{\percent}, $\uparrow$) of Faster R-CNN equipped with different defense methods against digital world transfer attacks.
    Each column corresponds to a texture crafted against the undefended detector or the corresponding defended detector. Ensemble denotes the ensemble attack, where the texture was jointly optimized against the undefended detector, the AT-defended detector and the FNC-defended detector. Each row corresponds to the detection performance of the undefended detector or the detector with a defense method on four adversarial textures. The best defense performance against each adversarial texture is marked in bold.}
    \label{tab:transfer-dig}
    \centering
    \begin{tabular}{cccccccccc}
        \toprule
Model & Undefended & AT \citep{li2023od-at} & FNC \citep{fnc-yu2021} & Ensemble \\
\midrule
Undefended & 0.03 & 4.40 & 1.94 & 0.19 \\
AT \citep{li2023od-at} & $\mathbf{62.65}$ & 5.97 & $\mathbf{69.12}$ & 11.08 \\
FNC \citep{fnc-yu2021} & 7.59 & 22.30 & 0.91 & 3.71 \\
LGS \citep{lgs-naseer2019} & 0.87 & 7.09 & 4.41 & 1.19 \\
IDBD \citep{epgf-zhou2020} & 1.72 & 5.05 & 4.20 & 0.76 \\
SAC \citep{sac-liu2022} & 0.43 & 4.67 & 2.23 & 0.60 \\
APE \citep{ape-kim2022} & 0.05 & 4.99 & 2.03 & 0.36 \\
UDF \citep{udf-yu2022} & 1.51 & 8.01 & 6.31 & 1.65 \\
Jedi \citep{jedi-tarchoun2023} & 5.32 & $\mathbf{24.18}$ & 8.04 & $\mathbf{17.24}$ \\
NAPGuard \citep{wu2024napguard} & 0.02 & 5.72 & 2.26 & 0.60 \\
        \bottomrule
    \end{tabular}
\end{table*}
We tested the performance of various adversarial patch defenses against texture-based attacks, and the results are shown in \cref{tab:dig-fr-3d}.
While random texture did not lower AP much, the adversarial texture broke all defense methods including the undefended model with APs all lower than \SI{5.32}{\percent}, except AT and FNC.
Since AT and FNC defense methods are hard to attack (AP $>\SI{7}{\percent}$) with the non-adaptive adversarial pattern optimized on the undefended Faster R-CNN model, we utilized the straightforward adaptive attack, \ie, optimizing the adversarial pattern on the detection model equipped with the defense method.
Even with adaptive attack, the strongest defense method is AT, with an AP of \SI{5.97}{\percent}.

The results of FCOS \citep{fcos} were similar as those of Faster R-CNN, as shown in \cref{tab:dig-fcos-3d}.
The texture-based attack broke all defenses with APs all lower than \SI{6.74}{\percent}. The strongest defense method is UDF with the highest AP. The conclusion is similar to that of Faster R-CNN detector.

We then evaluated the performance of adversarial defense methods against transfer attacks with adversarial textures optimized on the undefended Faster R-CNN detector, the detectors with AT or FNC defense, and an ensemble of the above three models, and the results are shown in \cref{tab:transfer-dig}.
The results show that the adversarially trained detector achieved an AP of \SI{62.65}{\percent} on the texture optimized on the undefended model, and an AP of \SI{69.12}{\percent} on the texture optimized on FNC model, indicating that AT succeeded in defending against the texture optimized on the undefended detector and the texture optimized on the FNC-defended detector.
Adaptively attacking AT model got an AP of \SI{5.97}{\percent},
but on the other hand deminished the adversarial effect against FNC defense, increasing AP to \SI{22.30}{\percent}.

\paragraph{Targeting an ensemble of defenses}
The 2nd to 4th columns of \cref{tab:transfer-dig} show that the adversarial textures crafted against the undefended Faster R-CNN, the AT model and the FNC-defended model were all defended by some defense method, with the APs of the respective best-performing defense on the texture above \SI{24.18}{\percent}.
Therefore, we optimized the texture on an ensemble of the undefended model, AT model and FNC-defended model,
denoted as \textit{Ensemble} in \cref{tab:transfer-dig}.
Not only did the \textit{Ensemble} texture get good adversarial effect against all models that were optimized on,
but it also transferred well to other defenses.
Overall, optimizing adversarial texture on an ensemble of defended models decreased the performances of all defended models to APs lower than \SI{17.24}{\percent}.

\begin{figure}[t]
    \centering
    \includegraphics[width=0.7\linewidth]{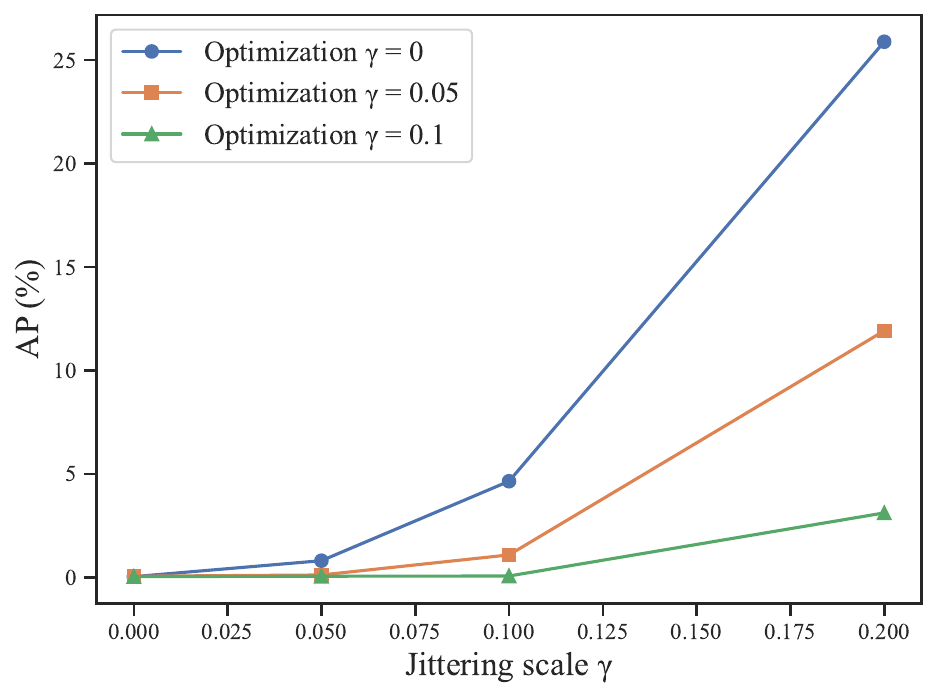}
    \caption{APs of Faster R-CNN versus jittering scale in the digital world for different jittering scales during texture optimization.
    $\gamma=0$ denotes no position jittering.
    }
    \label{fig:position-jitter}
\end{figure}

\paragraph{The effect of position jittering}
We optimized and evaluated the texture-based attack with various position jittering scales $\gamma$. The results are shown in \cref{fig:position-jitter}, with $\gamma=0$ indicating no position jittering.
When tested with jittering, the APs of the detector increased, especially when the textures optimized without jittering was applied.
This indicates the loss of adversarial effect when the position of the adversarial texture is applied with some error, which is inevitable in physical-world experiments because the shape of person varies.
The robustness of the attack to position jittering improved with higher values of $\gamma$ during optimization.
Therefore, applying position jittering let the attack resist to physical implementation errors and different body shapes of different individuals.
In this work, we fixed $\gamma=0.1$ when optimizing 3D textures.

\subsection{Physical-world evaluation}
\label{sec:phys-result}

\begin{figure}[t]
    \centering
    \includegraphics[width=0.9\linewidth]{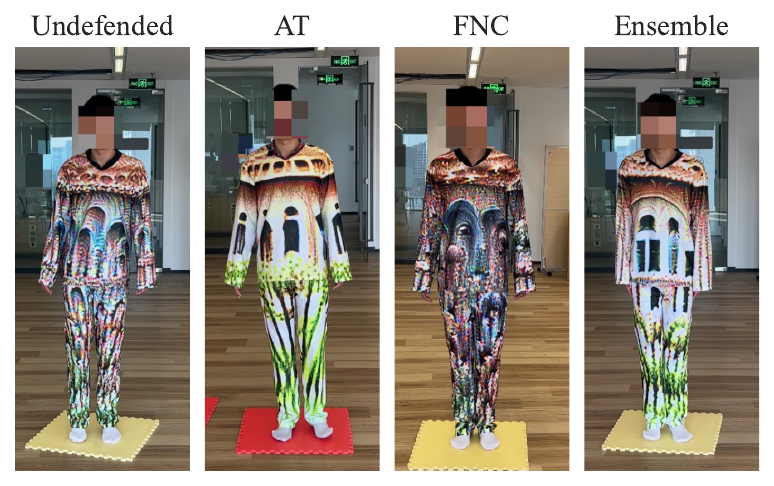}
    \caption{Visualization of different adversarial clothes produced in the physical world.
    }
    \label{fig:vis-clothes}
\end{figure}

\begin{table*}[t]
\small
\caption{ASRs (\si{\percent}, $\downarrow$) evaluated on Faster R-CNN equipped with different defense methods against physical world transfer attacks. Each column corresponds to a texture optimized on the detector equipped with the corresponding defense method. Ensemble denotes an ensemble of the undefended detector, the AT-defended detector and the FNC-defended detector. The best defense performance against each set of adversarial clothes is marked in bold.}
    \label{tab:transfer-phys}
    \centering
    \begin{tabular}{cccccccccc}
        \toprule
Model & Undefended & AT \citep{li2023od-at} & FNC \citep{fnc-yu2021} & Ensemble \\
\midrule
Undefended & 96.09 & 67.97 & 78.12 & 96.09 \\
AT \citep{li2023od-at} & $\mathbf{5.47}$ & 76.56 & $\mathbf{7.81}$ & $\mathbf{64.84}$ \\
FNC \citep{fnc-yu2021} & 95.31 & $\mathbf{46.88}$ & 99.22 & 98.44 \\
LGS \citep{lgs-naseer2019} & 28.12 & 78.91 & 13.28 & 82.81 \\
IDBD \citep{epgf-zhou2020} & 25.00 & 50.78 & 56.25 & 69.53 \\
SAC \citep{sac-liu2022} & 96.09 & 67.97 & 78.12 & 96.09 \\
APE \citep{ape-kim2022} & 92.19 & 65.62 & 75.00 & 94.53 \\
UDF \citep{udf-yu2022} & 72.66 & 71.09 & 33.59 & 81.25 \\
Jedi \citep{jedi-tarchoun2023} & 78.91 & 83.59 & 96.88 & 87.50 \\
NAPGuard \citep{wu2024napguard} & 96.09 & 67.97 & 78.12 & 96.09 \\
        \bottomrule
    \end{tabular}
\end{table*}

For physical implementation, the textures optimized for different defenses were printed on fabric and subsequently tailored into shirts and trousers.
See \cref{fig:vis-clothes} for the visualization of four sets of clothes, whose textures have been optimized on different defended models.
To assess the real-world effectiveness, we measured the ASRs from various viewing angles by capturing a video of a person turning circles.
Frames of different angles were evenly extracted and ASRs on these frames were computed.
\Cref{tab:transfer-phys} shows the physical world ASRs evaluated on Faster R-CNN equipped with different defense methods, against clothing textures optimized on the undefended detector, AT-defended detector, FNC-defended detector, and an ensemble of the above three models.
The results are mostly consistent with those in the digital world (\cref{tab:transfer-dig}).
The adversarially trained detector successfully defended non-adaptive textures and FNC-adaptive textures, with only \SI{5.47}{\percent} and \SI{7.81}{\percent} ASR, respectively.
However, AT was defeated with AT-adaptive texture and the texture optimized on an ensemble of defended models, with ASRs of \SI{76.56}{\percent} and \SI{64.84}{\percent}, respectively.
LGS was also defeated by this two kinds of textures, with ASRs of \SI{78.91}{\percent} and \SI{82.81}{\percent}.
IDBD performed well in defending against all three textures optimized on a single model, but the ASR increased to \SI{69.53}{\percent} when defending against the texture optimized on an ensemble of models.
FNC worked well in clipping adversarial features on the AT-adaptive texture, but failed to defend against other textures, with ASRs all above \SI{95.31}{\percent}.
Jedi, although performed well in the digital world in defending against the adversarial textures with an AP of \SI{17.24}{\percent} on the Ensemble one, the corresponding physical world ASR evaluated on Jedi was \SI{87.50}{\percent}, ranked only fifth among the nine defense methods evaluated, behind AT, IDBD, UDF and LGS. The results indicate that the good performance of Jedi in the digital world relies on the less realistic 3D rendering result.

When evaluated against the texture optimized on an ensemble of three defense models, all nine defenses failed to defend against the adversarial texture, with ASRs against them all above \SI{64.84}{\percent}. The ASR against the undefended model was \SI{96.09}{\percent}.

\begin{figure}[t]
    \centering
    \includegraphics[width=0.9\linewidth]{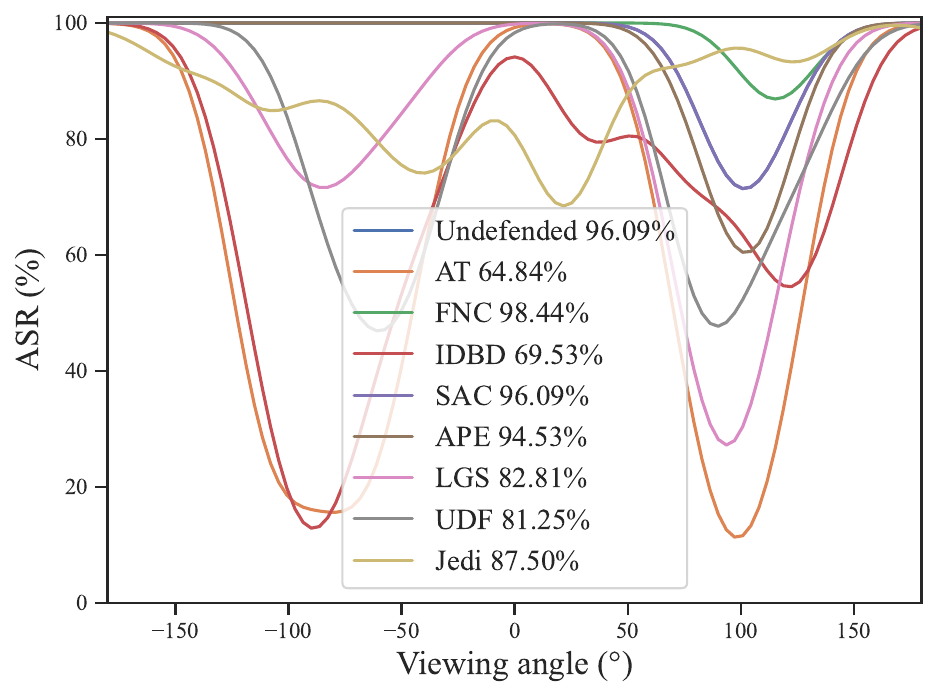}
    \caption{ASRs evaluated on the defended detectors at a distance of \SI{4}{m} from different viewing angles on the Ensemble adversarial clothes in the physical world. The person faces the camera when the viewing angle equals \SI{0}{\degree}. 
    In the legend, we show the ASRs averaged over viewing angles.
    }
    \label{fig:angles}
\end{figure}

\paragraph{Different viewing angles}
\Cref{fig:angles} presents ASRs evaluated on the defended detectors against the Ensemble adversarial clothes in the physical world under different viewing angles from \SI{-180}{\degree} to \SI{180}{\degree}.
We observed that viewing angles of $\SI{\pm 90}{\degree}$ (one side of person is facing the camera) were where the defenses performed best, as most defense methods got lower ASRs.
This is probably because the side of person has smaller area, and only has a small area of adversarial pattern captured in the camera.

It is worth mentioning that in the physical world, most patch-based attacks \citep{thys2019fooling,hu2021nat-patch,xu2020adv-tshirt} evaluated ASRs from the front view of the person. This is primarily because the patch needs to be fully facing the camera to effectively execute the attack. Therefore, in addition to examining ASRs from all viewing angles, we specifically focused on the viewing angle of \SI{0}{\degree}. Remarkably, regardless of the defense employed, the front view consistently yielded ASRs surpassing \SI{80}{\percent}.

\begin{figure}[t]
    \centering
    \includegraphics[width=0.9\linewidth]{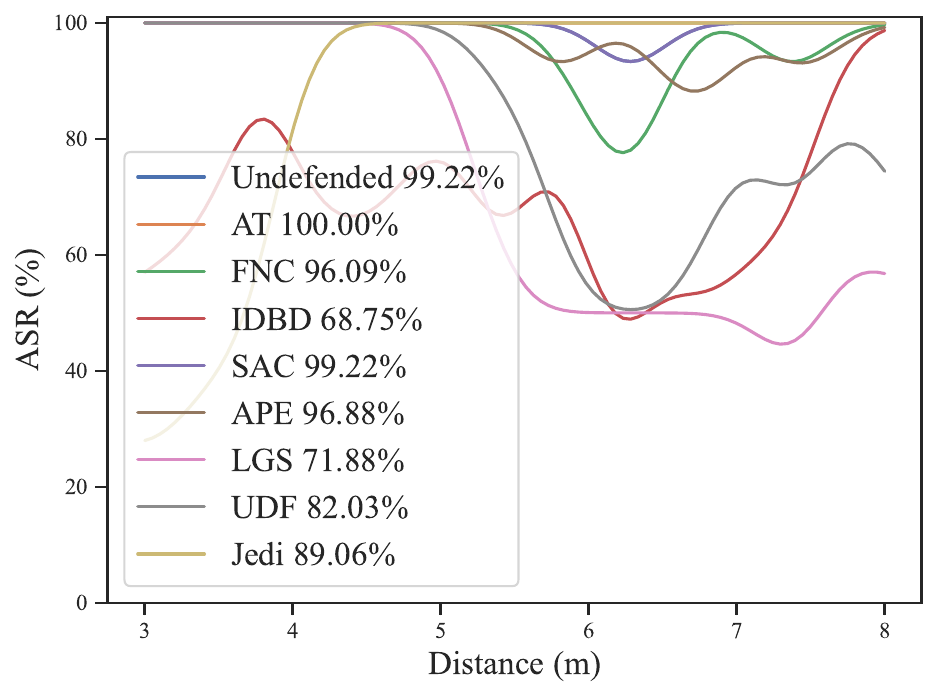}
    \caption{ASRs evaluated on the defended detectors at different distances with the Ensemble adversarial clothes in the physical world.
    In the legend, we show the ASRs averaged over distances.
    }
    \label{fig:distance}
\end{figure}

\paragraph{Different distances}
We tested the performance of defenses against the adversarial clothes when the person is of different distances away from the camera.
We captured a video of a person moving close to the camera facing it, and moving away from the camera facing opposite to it, thus fixing the viewing angle to \SI{0}{\degree} and \SI{180}{\degree}.
ASRs at different distances are shown in \cref{fig:distance}.
The ASRs evaluated on the undefended model were steadily high as distance changed, with average ASRs across distances achieving \SI{99.22}{\percent}.
A significant drop of ASR when distance increased appeared on LGS and UDF defenses, especially when the distance exceeded \SI{6}{\meter}.
IDBD and Jedi, the two entropy-based defenses, performed better when distance decreased.

\begin{figure}[t]
    \centering
    \includegraphics[width=0.9\linewidth]{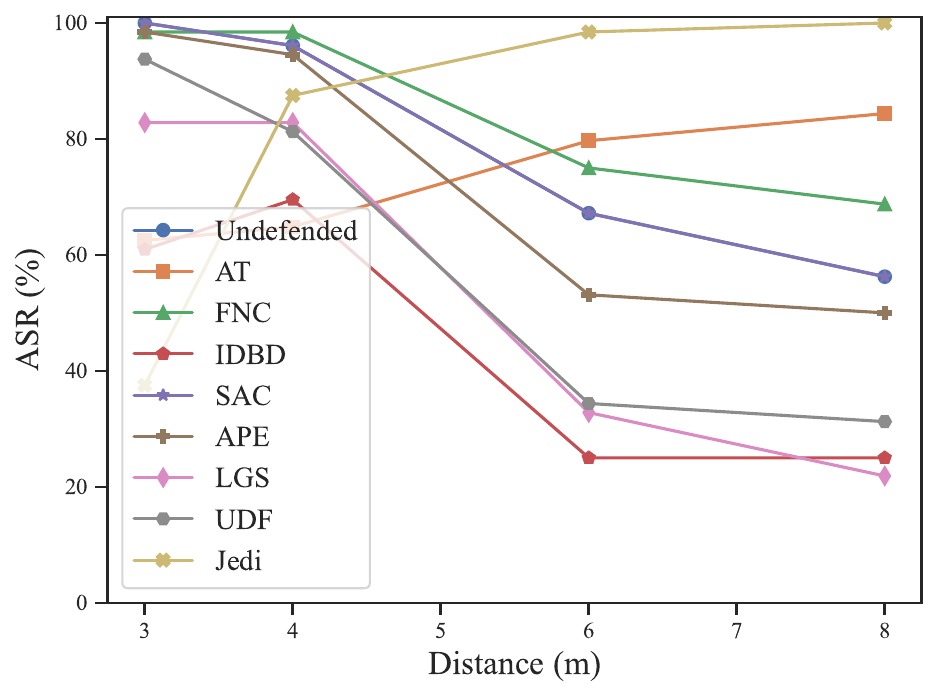}
    \caption{ASRs evaluated on the defended detectors at different distances averaged over all viewing angles from \SIrange{-180}{180}{\degree} with the Ensemble adversarial clothes in the physical world.
    Distance of \SI{4}{\meter} is where prior physical experiments were conducted.
    }
    \label{fig:distance-discrete}
\end{figure}

\paragraph{Different distances for all viewing angles}
\Cref{fig:distance-discrete} presents ASRs averaged over all viewing angles (\SIrange{-180}{180}{\degree}) at different distances.
With the distance increasing, the ASRs of all defenses declined except AT and Jedi. 
The worse performance of AT model when distance increased was likely attributed to the diminished performance of the AT model in detecting smaller objects, as detailed by \citet{li2023od-at}.
The performance of Jedi defense when distance changed was consistent with that evaluated on only two view angles (see \cref{fig:distance}).
Nevertheless, the ASRs against all defenses were still over \SI{20}{\percent} across all distances.

\section{Conclusion}
\label{sec:conclusion}

In this study, we show that a single set of adversarial clothes broke nine defense methods in real-world scenario.
Motivated by the finding that enlarged patch broke a typical size-independent defense method, we evaluated nine different defense methods against adversarial clothes.
All defense methods had poor performance against adversarial clothes in both the digital world and the physical world.
Moreover, we created a single set of adversarial clothes by optimizing the adversarial texture on an ensemble of three defended models.
The adversarial clothes achieved an ASR of \SI{96.06}{\percent} on the undefended model, and broke nine defended models with ASRs over \SI{64.84}{\percent} in the physical world.
More detailed analyses on viewing angles show that the defense methods worked better with either side of the person facing the camera.
Furthermore, different defense methods performed differently as distance between the person and the camera varied, but the overall defense performance was still sub-optimal.

This paper reveals that SOTA defense methods are still commonly vulnerable to physical-world adversarial examples when confronted with texture-based adversarial attacks.
Therefore, there is an urgent need for future adversarial defenses to consider a broader range of attacks, at least including adversarial clothes.

\section*{Acknowledgement}
This work was supported by the National Natural Science Foundation of China under grant U2341228.

\bibliographystyle{model2-names}
\bibliography{refs}

\end{document}